\documentclass[letterpaper, 10 pt, conference]{ieeeconf}  

\IEEEoverridecommandlockouts                              

\usepackage{cite}
\usepackage{amsmath,amssymb,amsfonts}
\usepackage{algorithmic}
\usepackage{graphicx}
\usepackage{textcomp}
\usepackage{xcolor}
\usepackage{hyperref}
\usepackage{cuted} 
\usepackage{capt-of}
\usepackage{booktabs}
\usepackage{nicematrix}
\usepackage{float}
\newcommand\blfootnote[1]{%
  \begingroup
  \renewcommand\thefootnote{}\footnote{#1}%
  \addtocounter{footnote}{-1}%
  \endgroup
}
\begin{document}

\thispagestyle{empty}
\pagestyle{empty}

\begin{strip}
    \centering

    {\LARGE\bfseries
    A Generalizable Approach to Safety-Aware Robotic Behavior Planning\\
    with an Explicit Constraint Space
    \par}

    \vspace{0.6cm}

    {\large
    Xianxu Wu$^{*}$, 
    Shang-Ching Liu$^{*}$, 
    Zhaocong Liang, 
    Hengxiang Chen, 
    Yujian Ma,\\
    Shenwen Deng, 
    Qiang Li$^{\dagger}$, 
    Yung-Hui Li, 
    and Jianwei Zhang
    \par}

    \vspace{0.45cm}
    \includegraphics[width=0.9\textwidth]{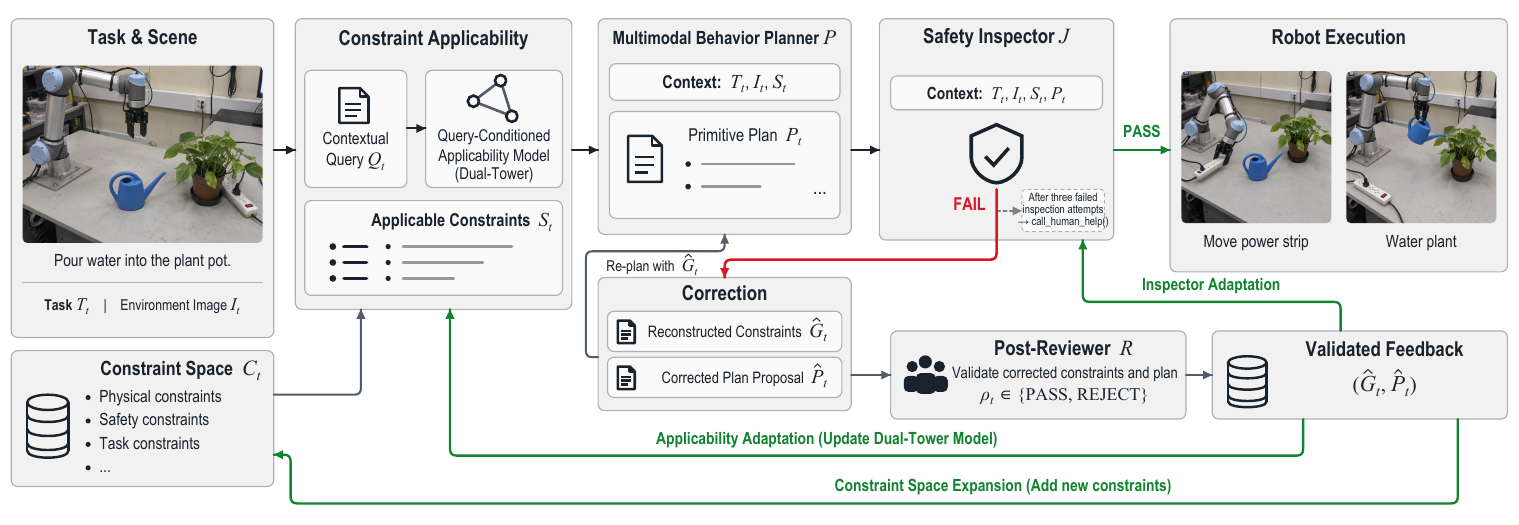}

    \vspace{-0.15cm}

    \captionof{figure}{
\textbf{Closed-loop architecture for safety-aware behavior planning with adaptive constraint knowledge.}
Task--scene-relevant constraints are selected from the Constraint Space and provided to a multimodal behavior planner. Unsafe plans trigger Inspector-guided constraint reconstruction and re-planning. Post-reviewed feedback updates the applicability model, adapts the Safety Inspector through LoRA, and expands the Constraint Space. Only plans receiving \texttt{PASS} are executed.
}
    \label{fig:1}

\end{strip}
\blfootnote{$^{*}$Equal contribution. $^{\dagger}$Corresponding author.}
\begin{abstract}
Large language and vision-language models have substantially advanced robotic behavior planning from natural-language instructions and visual observations. However, safe manipulation in tabletop environments remains challenging because the safety constraints required for a task depend on the surrounding scene and may not be explicitly specified in the instruction. We propose a safety-aware behavior planning framework based on an explicit Constraint Space that stores reusable textual safety rules. Rather than retrieving rules solely by semantic similarity, we formulate constraint selection as task- and environment-conditioned applicability learning. A lightweight query-conditioned dual-tower model identifies the constraints applicable to the current context, which are then provided together with the original task and image to a multimodal behavior planner. To handle incorrect constraint selections and previously unseen hazards, we introduce a multimodal Safety Inspector that evaluates generated primitive plans, reconstructs the complete applicable constraint set, and provides corrective planning feedback. Validated inspection feedback is further used for test-time adaptation of constraint applicability and parameter-efficient adaptation of the Safety Inspector through LoRA, while newly identified constraints are incorporated into the Constraint Space. The resulting framework integrates explicit safety knowledge representation, context-aware constraint selection, corrective safety inspection, and test-time knowledge adaptation for safety-aware robotic behavior planning.
\end{abstract}

\section{INTRODUCTION}\label{s:1}

Large language models (LLMs) and vision-language models (VLMs) have significantly advanced robotic behavior planning by enabling robots to interpret natural-language instructions, understand visual scenes, and generate long-horizon action sequences~\cite{RN4,RN5,RN6}. However, completing a task does not necessarily imply completing it safely. In tabletop manipulation, the same instruction may involve different safety requirements depending on the surrounding environment. For example, manipulating a liquid near electronic devices or operating a sharp object near a human requires safety considerations that may not be explicitly stated in the task instruction. Therefore, a safety-aware robot must not only understand what task to perform, but also identify the safety constraints imposed by the current task and environment~\cite{RN1}.

To make such safety knowledge explicit and reusable, we maintain a \emph{Constraint Space} consisting of textual safety rules. This representation separates safety knowledge from the planner itself and allows the same constraints to be reused across different tasks and scenes. However, maintaining a rule library introduces another challenge: only a small subset of constraints is typically applicable to a particular task and environment. Selecting rules merely according to semantic similarity does not necessarily reflect whether a constraint should actually be enforced in the current scene. The key problem is therefore to determine the \emph{applicability} of each constraint conditioned jointly on the task and the environment.

We address this problem with a lightweight query-conditioned constraint applicability model. Given a natural-language task and a visual observation, the environment is first represented in textual form and combined with the task to form a contextual query. A dual-tower model then evaluates the constraints in the Constraint Space and selects an applicable subset, which is provided together with the original task and image to a multimodal behavior planner. In this way, safety constraints are explicitly selected before planning rather than being left entirely to the implicit safety reasoning of the planner.

A fixed Constraint Space and a fixed applicability model, however, cannot cover every safety situation encountered at test time. The selected constraint set may contain irrelevant rules, omit safety-critical ones, or fail to represent a previously unseen hazard. We therefore introduce a multimodal Safety Inspector that examines the generated primitive plan and, when necessary, jointly produces a corrected plan proposal and reconstructs the complete set of constraints applicable to the current task and environment. The reconstructed constraints are returned to the planner for corrective re-planning. After validation by a post-reviewer, Inspector-generated feedback is further used for test-time adaptation of both the constraint-applicability model and the Safety Inspector, with the latter adapted through LoRA. Newly identified constraints are also incorporated into the Constraint Space. In this way, the framework can continuously improve constraint selection, safety inspection, and the explicit safety knowledge available for future planning.

Our contributions are threefold:
\begin{itemize}
\item \textbf{Query-Conditioned Constraint Applicability Learning.}
We formulate safety-constraint selection as task- and environment-conditioned applicability learning and introduce a lightweight dual-tower model in which each contextual query induces a separating boundary over learned constraint representations.

\item \textbf{Multimodal Safety Inspection with Constraint Reconstruction.}
We introduce a multimodal Safety Inspector that goes beyond binary safety verification by jointly producing a corrected primitive-plan proposal and reconstructing the complete set of constraints applicable to the current task and environment.

\item \textbf{Inspector-Guided Test-Time Adaptation and Knowledge Expansion.}
We use validated Inspector feedback to adapt both the lightweight constraint-applicability model and the Safety Inspector through parameter-efficient LoRA updates, while newly identified constraints are incorporated into the Constraint Space for explicit safety-knowledge expansion.
\end{itemize}

\section{RELATED WORK}\label{s:2}

This section reviews three lines of research relevant to our work. We first discuss safety-aware robotic manipulation, focusing on scene-dependent hazards and safety constraints in manipulation tasks. We then review LLM-based task planning methods that translate natural-language instructions into structured and executable robot behaviors. Finally, we discuss test-time adaptation methods that enable models to improve from deployment-time feedback.

\subsection{Safety-Aware Robotic Manipulation}\label{s:2.1}

As large language and vision-language models are increasingly used for robotic manipulation, ensuring that generated behaviors remain safe under the current task and environment has become an important problem. Beyond geometric feasibility and collision avoidance, manipulation safety may also depend on semantic relationships among objects in the scene. Recent work on semantically safe robot manipulation has studied such scene-dependent safety requirements in tabletop environments, where interactions such as moving a liquid-filled container near electronic devices may introduce hazards that are not fully captured by geometric constraints alone. This work combines semantic scene understanding with language-model reasoning to identify semantically unsafe conditions and map them to motion-level safeguards~\cite{brunke2024semanticallysafe}.

Safety-aware reasoning has also been investigated at the behavior-planning level. Safety-as-Policy requires robots to consider potential hazards and their consequences while completing manipulation instructions, enabling safety considerations to influence behavior generation rather than focusing solely on task completion~\cite{RN15}. Complementary work has explored explicit enforcement of externally specified safety requirements. Safety Chip translates natural-language safety constraints into Linear Temporal Logic specifications and uses them to monitor LLM-generated actions, reason about constraint violations, and prevent unsafe actions from being executed~\cite{yang2023safetychip}.

These studies demonstrate the importance of both scene-dependent hazard reasoning and explicit safety constraints for responsible robotic manipulation. However, when safety knowledge is maintained as a reusable collection of constraints, a key challenge is determining which constraints should apply to the current task and environment, and how this constraint knowledge should evolve as new safety situations are encountered.
\subsection{LLM-Based Robotic Task Planning}\label{s:2.2}

Large language models have enabled robots to translate natural-language instructions into structured task plans and executable behaviors. ProgPrompt generates situated robot task plans by providing an LLM with programmatic descriptions of the environment, including available objects and actions~\cite{RN13}. Code as Policies further treats language models as code-writing policies, generating executable robot programs that compose predefined perception and control APIs~\cite{RN10}. SayCan combines the semantic reasoning capability of language models with the affordances of pretrained robot skills, allowing the planner to select actions that are both relevant to the instruction and feasible for the robot to execute~\cite{RN3}.

These methods demonstrate the effectiveness of language models for high-level robotic task planning and behavior generation. However, task relevance and physical feasibility alone do not guarantee safe behavior in tabletop environments, where additional safety requirements may arise from the interaction between the task and the surrounding scene. This highlights the need for robotic planners that can explicitly account for scene-dependent safety knowledge during behavior generation.

\subsection{Test-Time Adaptation}\label{s:2.3}

Test-time adaptation enables a model to update its behavior during deployment using information obtained from incoming test data. Test-time training has shown that self-supervised signals available at inference time can be used to improve model robustness under distribution shift~\cite{sun2020test}. For large pretrained models, parameter-efficient approaches such as LoRA~\cite{hu2022lora} provide a lightweight mechanism for adaptation by updating only a small set of additional parameters while keeping the underlying backbone frozen.

Our framework uses deployment-time feedback differently from conventional self-supervised test-time adaptation. The adaptation signal is generated by the closed-loop Safety Inspector and validated before being incorporated into the system. The validated feedback is used to update the lightweight constraint-applicability model, adapt the Safety Inspector through LoRA, and expand the explicit Constraint Space with newly identified safety rules. This allows test-time improvement to occur jointly in constraint selection, safety inspection, and reusable safety-knowledge representation.

\section{Methodology}\label{s:3}
\subsection{Problem Formulation and System Overview}\label{s:3.1}

Given a natural-language task \(T_t\) and an environment image \(I_t\), our goal is to generate a primitive action sequence \(P_t\) that accomplishes the task while satisfying the safety requirements induced by the current task--environment context. We maintain an external Constraint Space \(C_t\) containing reusable textual safety knowledge, from which an applicable constraint set \(S_t \subseteq C_t\) is selected for the current input. A multimodal behavior planner then generates
\[
P_t=\mathcal{P}(T_t,I_t,S_t).
\]

Before execution, the generated plan is evaluated by a multimodal Safety Inspector. A plan is executed only after receiving a \texttt{PASS} decision. Otherwise, the Inspector jointly reconstructs the complete applicable constraint set \(\hat{G}_t\) and produces a corrected primitive-plan proposal \(\hat{P}_t\). The reconstructed constraint set, rather than the Inspector-proposed plan, is returned to the planner for corrective re-planning:
\[
P_t'=\mathcal{P}(T_t,I_t,\hat{G}_t).
\]

The resulting plan is inspected again, and the planning--inspection loop continues until a \texttt{PASS} decision is obtained. After three failed inspection attempts, the system invokes \texttt{call\_human\_help()} as a safety fallback.

Validated Inspector feedback is further reused during deployment for test-time adaptation. Specifically, it provides supervision for updating the lightweight constraint-applicability model, adapting the Safety Inspector through LoRA, and expanding the explicit Constraint Space with newly identified safety rules.
\subsection{Query-Conditioned Constraint Applicability Learning}\label{s:3.3}

Given the Constraint Space \(C_t\) introduced above, let \(c_i \in C_t\) denote the \(i\)-th textual safety constraint. The first stage of our framework determines which stored constraints should apply to the current task and environment. A constraint can be semantically related to a task while still being unnecessary under the current scene configuration. We therefore formulate constraint selection as a task--environment-conditioned applicability learning problem rather than relying solely on semantic similarity.

Given a natural-language task \(T_t\) and a 2D environment image \(I_t\), a vision-language model first converts the visual observation into a textual environment description
\[
D_t=\mathcal{V}(I_t).
\]
The task and environment description are then concatenated to form a contextual query
\[
Q_t=T_t\oplus D_t.
\]
A frozen BGE-M3 encoder~\cite{chen2024m3embedding} \(E(\cdot)\) is used to encode both the contextual query and each constraint:
\[
q_t=E(Q_t)\in\mathbb{R}^{1024},
\qquad
r_i=E(c_i)\in\mathbb{R}^{1024}.
\]
The encoder remains frozen during both offline training and test-time adaptation.

Rather than directly comparing \(q_t\) and \(r_i\) using semantic similarity, we introduce a lightweight dual-tower constraint applicability model. The rule tower maps each constraint embedding into a learned constraint representation:
\[
x_i=f_r(r_i)\in\mathbb{R}^{512},
\]
where \(f_r\) is an MLP with dimensions \(1024\rightarrow1024\rightarrow512\).

The query tower predicts a query-conditioned separating hyperplane. Specifically,
\[
f_q(q_t)=
\begin{bmatrix}
w_t\\
b_t
\end{bmatrix}
\in\mathbb{R}^{513},
\]
where \(w_t\in\mathbb{R}^{512}\) and \(b_t\in\mathbb{R}\), and \(f_q\) is an MLP with dimensions \(1024\rightarrow1024\rightarrow513\). Each task--environment query therefore defines its own separating hyperplane
\[
w_t^\top x+b_t=0
\]
over the learned constraint representations.

For each constraint \(c_i\), we compute its normalized signed distance to the query-conditioned hyperplane as
\[
d_{t,i}=
\frac{x_i^\top w_t+b_t}
{\|w_t\|_2}.
\]
A positive \(d_{t,i}\) places the constraint on the applicable side of the query-conditioned boundary, whereas a negative value places it on the opposite side. We then apply a sigmoid transformation
\[
s_{t,i}=\sigma(d_{t,i}),
\]
where \(s_{t,i}\in(0,1)\) is the constraint applicability score. Since the sigmoid function is monotonic, this transformation preserves the ordering induced by the signed distance while mapping it to a normalized range suitable for binary classification and constraint selection.

During offline training, each task--environment query is paired with manually annotated applicable constraints. Let
\[
y_{t,i}\in\{0,1\}
\]
indicate whether \(c_i\) is applicable to the current query. Because the number of non-applicable constraints is substantially larger than the number of applicable constraints, we apply a positive weight to the binary cross-entropy objective. Let \(N_t^{+}\) and \(N_t^{-}\) denote the numbers of applicable and non-applicable constraints for query \(Q_t\), respectively. The positive weight is defined as
\[
\alpha_t=\frac{N_t^{-}}{N_t^{+}},
\]
and the offline training objective is
\[
\mathcal{L}_{\mathrm{offline}}
=
-\sum_i
\left[
\alpha_t y_{t,i}\log s_{t,i}
+
(1-y_{t,i})\log(1-s_{t,i})
\right].
\]
The weighting is used only to compensate for class imbalance during offline training. Only the lightweight query and rule towers are optimized, while the BGE-M3 encoder remains fixed.

At inference time, constraint applicability is determined directly by the sign of the normalized signed distance. Specifically, the applicable constraint set is defined as
\[
S_t
=
\left\{
c_i\in C_t
\mid
d_{t,i}>0
\right\}.
\]
Thus, constraints on the positive side of the query-conditioned hyperplane are selected as applicable, while those on the opposite side are excluded. The sigmoid transformation \(s_{t,i}=\sigma(d_{t,i})\) provides a monotonic normalization of the signed distance for binary-cross-entropy training, whereas the inference decision is determined directly by the sign of \(d_{t,i}\).

The textual description \(D_t\) is used to construct the applicability query, whereas the original image \(I_t\) is retained for behavior planning. After obtaining \(S_t\), the multimodal planner receives \(T_t\), \(I_t\), and \(S_t\) to generate the corresponding primitive action sequence.
\subsection{Multimodal Safety Inspection with Constraint Reconstruction}\label{s:3.3}

The constraint applicability model determines which safety constraints are provided to the behavior planner, but an incorrect or incomplete selection may still lead to an unsafe or inadequate plan. We therefore introduce a multimodal Safety Inspector that evaluates the generated primitive action sequence before execution and provides structured corrective feedback when necessary.

Given the current task \(T_t\), the original environment image \(I_t\), the selected constraint set \(S_t\), and the planner-generated primitive action sequence \(P_t\), we define the inspection state as
\[
X_t=(T_t,I_t,S_t,P_t).
\]

The Inspector jointly reasons over the visual environment, the applied constraints, and the proposed behavior. Its output takes one of two forms:
\[
\mathcal{I}(X_t)=
\begin{cases}
\texttt{PASS}, & \text{if the current plan is safe},\\
(\texttt{FAIL},\hat{G}_t,\hat{P}_t), & \text{otherwise}.
\end{cases}
\]

Here, \(\hat{G}_t\) denotes the reconstructed complete applicable constraint set and \(\hat{P}_t\) denotes an Inspector-generated corrected primitive-plan proposal. Constraint reconstruction considers the complete safety requirements of the current context: applicable constraints already contained in \(S_t\) are retained, inappropriate selected constraints are removed, and missing constraints are added. Thus, \(\hat{G}_t\) represents a corrected and complete applicable constraint set rather than a single newly generated rule.

The reconstructed constraints are then fed back to the multimodal behavior planner for re-planning:
\[
P_t^{(k+1)}
=
\mathcal{P}
\left(
T_t,I_t,\hat{G}_t^{(k)}
\right),
\]
where \(k\) denotes the current inspection attempt. The resulting primitive plan is subsequently evaluated again using
\[
X_t^{(k+1)}
=
\left(
T_t,I_t,\hat{G}_t^{(k)},P_t^{(k+1)}
\right).
\]

This planning--inspection--replanning loop continues until the Inspector returns \texttt{PASS}. After three failed inspection attempts, the system invokes
\[
\texttt{call\_human\_help()}
\]
as the final safety fallback.

The corrected plan proposal \(\hat{P}_t\) is not directly executed. Instead, together with \(\hat{G}_t\), it provides structured corrective information that is validated and subsequently reused for test-time adaptation.
\subsection{Inspector-Guided Test-Time Adaptation and Knowledge Expansion}
\label{s:3.4}

The Safety Inspector not only provides corrective feedback for unsafe planning results, but also produces structured supervision that can be reused during deployment for test-time adaptation~\cite{sun2020test}. To reduce the risk of propagating erroneous self-generated feedback, Inspector-generated corrections are first validated by a prompt-constrained post-reviewer before entering the adaptation loop.

Given an Inspector correction \((\hat{G}_t,\hat{P}_t)\), the post-reviewer evaluates the corrected result under the current task and visual environment:
\[
\rho_t
=
\mathcal{R}(T_t,I_t,\hat{G}_t,\hat{P}_t),
\]
where
\[
\rho_t\in\{\texttt{PASS},\texttt{REJECT}\}.
\]
Only corrections receiving \texttt{PASS} are used for subsequent adaptation and knowledge update.

\textbf{Constraint-applicability adaptation.}
For an approved correction, the reconstructed constraint set \(\hat{G}_t\) provides supervision for updating the constraint-applicability model. Let \(S_t\) denote the constraint set originally selected for the current query. Positive supervision is constructed as
\[
D_t^{+}
=
\{(q_t,r_i,1)\mid c_i\in\hat{G}_t\},
\]
while negative supervision is constructed from constraints that were initially selected but removed by the Inspector:
\[
D_t^{-}
=
\{(q_t,r_i,0)\mid c_i\in S_t\setminus\hat{G}_t\}.
\]
The resulting adaptation set is
\[
D_t=D_t^{+}\cup D_t^{-}.
\]
Constraints appearing in neither \(S_t\) nor \(\hat{G}_t\) remain unlabeled and are not treated as negative samples.

Unlike offline training, test-time adaptation uses standard binary cross-entropy without class weighting. We write
\[
\ell(y,s)
=
y\log s+(1-y)\log(1-s),
\]
and define
\[
\mathcal{L}_{\mathrm{app}}
=
-\sum_{(q_t,r_i,y_i)\in D_t}
\ell(y_i,s_{t,i}).
\]
The lightweight dual-tower parameters are then updated by
\[
\theta_{t+1}
=
\theta_t
-
\eta_{\mathrm{app}}
\nabla_{\theta}\mathcal{L}_{\mathrm{app}},
\]
while the BGE-M3 encoder remains frozen.

\textbf{Safety Inspector adaptation.}
The Safety Inspector is additionally adapted during deployment using LoRA~\cite{hu2022lora}, while the underlying VLM backbone remains frozen. The adaptation data preserve the same input--output format used by the Inspector during inference.

Recall that the Inspector input is
\[
X_t=(T_t,I_t,S_t,P_t).
\]
For a planning state that is verified to satisfy the current task and safety requirements, a positive training sample is constructed as
\[
(X_t,\texttt{PASS}).
\]

For a failed planning state, the target contains both the \texttt{FAIL} decision and its corrective output. To simplify the notation, we define the correction produced at inspection round \(k\) as
\[
R_t^{(k)}
=
(\hat{G}_t^{(k)},\hat{P}_t^{(k)}).
\]

Consider a case in which the initial planner output requires two rounds of correction. The first inspection produces
\[
X_t^{(1)}
\xrightarrow{\mathcal{I}}
(\texttt{FAIL},R_t^{(1)}).
\]
The reconstructed constraints are then returned to the planner:
\[
P_t^{(2)}
=
\mathcal{P}
(T_t,I_t,\hat{G}_t^{(1)}),
\]
and the replanned result is inspected again:
\[
X_t^{(2)}
\xrightarrow{\mathcal{I}}
(\texttt{FAIL},R_t^{(2)}).
\]

The existence of the second correction indicates that additional corrective information is required beyond the first inspection result. Rather than training the Inspector to distribute this correction across multiple rounds, we directly associate the later corrective output with the initial inspection input and construct the failure sample
\[
\bigl(
X_t^{(1)},
(\texttt{FAIL},R_t^{(2)})
\bigr).
\]

In this way, corrective information discovered in a later inspection round is used to supervise the Inspector on the original failed planning state, encouraging it to produce a more complete correction in a single inspection step.

Let \(D_t^{\mathrm{ins}}\) denote the resulting validated Inspector adaptation samples. The LoRA parameters \(\phi\) are optimized using the standard autoregressive training objective:
\[
\phi_{t+1}
=
\phi_t
-
\eta_{\mathrm{ins}}
\nabla_{\phi}\mathcal{L}_{\mathrm{ins}},
\]
while the underlying VLM backbone remains frozen.

\textbf{Constraint Space expansion.}
Validated Inspector corrections can also expand the explicit safety knowledge stored in the Constraint Space. We define newly identified constraints as
\[
\Delta C_t
=
\hat{G}_t\setminus C_t,
\]
and update the Constraint Space as
\[
C_{t+1}
=
C_t\cup\Delta C_t.
\]
The newly added constraints then become available for applicability estimation in subsequent planning episodes.

Consequently, validated Inspector feedback supports three complementary forms of test-time improvement: parameter adaptation of the lightweight constraint-applicability model, parameter-efficient adaptation of the Safety Inspector, and non-parametric expansion of the explicit Constraint Space. Together, these mechanisms improve constraint selection, corrective safety inspection, and reusable safety knowledge during deployment.
\subsection{Implementation and Primitive Grounding Details}
\label{s:3.5}

The behavior planner represents each task plan as a sequence of predefined manipulation primitives. To ground these high-level primitives into executable robot behaviors, we employ a perception-and-grounding pipeline that connects symbolic primitive descriptions with task-relevant objects, object parts, and spatial relationships in the physical scene.

For each primitive, Grounding DINO~\cite{liu2024groundingdino} is first used to localize task-relevant objects or parts from language-conditioned visual queries, and SAM2~\cite{ravi2024sam2} is used to obtain their corresponding visual masks.

Given the grounded objects and parts, we use CoPa~\cite{RN22} to formulate the spatial constraints required by the current primitive.

For the vision-language components, GPT-5.5~\cite{openai2026gpt55} is used for environment-description generation, multimodal behavior planning, and post-review. The Safety Inspector backbone is configured separately according to the corresponding experiment, while its adaptation procedure is described in Sec.~III-D.

For safety-critical execution, each newly generated primitive sequence must receive a \texttt{PASS} decision from the Safety Inspector before being sent to the grounding and execution pipeline. If three consecutive planning attempts fail inspection, the system invokes \texttt{call\_human\_help()} as a safety fallback.

\section{EXPERIMENTS}\label{s:4}

We evaluate the framework on tabletop scenes spanning home, kitchen, office, and laboratory settings. The experiments cover constraint applicability under distribution shift, Safety Inspector performance, sequential test-time adaptation, and representative real-world robot manipulation.

\subsection{Constraint Applicability Evaluation}\label{s:4.1}

We evaluate whether the query-conditioned applicability model can transfer learned constraint applicability to unseen task--scene contexts more effectively than semantic retrieval. The model is trained on 50 original scenes with Constraint Space \(C_1\) and evaluated on 50 newly constructed scenes with \(C_2\); evaluation task instructions and scene descriptions are unseen during training, although \(C_2\) may contain rules also present in \(C_1\). Each test scene is paired with three task configurations (\emph{single}, \emph{double}, and \emph{multi}), resulting in 150 held-out task--scene queries. The benchmark defines a unified Constraint Space containing 93 candidate safety rules.

For each held-out query, our applicability model selects a dynamically sized constraint set \(S_t^{\mathrm{ours}}\) according to the signed-distance decision criterion defined in Sec.~III-B. Let
\[
n_t=\left|S_t^{\mathrm{ours}}\right|.
\]
As the semantic-retrieval baseline, we rank all rules in the same Constraint Space using cosine similarity between the contextual query and rule embeddings produced by BGE-M3~\cite{chen2024m3embedding}, and retrieve the top-\(n_t\) rules. Thus, both methods return the same number of constraints for each query, allowing the comparison to focus on which method better identifies truly applicable rules.

We compare the selected set \(S_t\) with the manually annotated ground-truth set \(G_t^\ast\) using Recall, Precision, F1 score, and Exact Match Rate (EMR). Recall and Precision are defined as
\[
\mathrm{Recall}_t
=
\frac{|S_t\cap G_t^\ast|}{|G_t^\ast|},
\qquad
\mathrm{Precision}_t
=
\frac{|S_t\cap G_t^\ast|}{|S_t|}.
\]
Recall is treated as the primary safety-oriented metric because missing an applicable constraint may leave a downstream hazard unaddressed, while Precision reflects the amount of irrelevant constraint information introduced into planning. F1 measures the balance between the two. EMR further measures the fraction of queries for which the selected constraint set exactly matches the complete ground-truth set:
\[
\mathrm{EMR}
=
\frac{1}{N}
\sum_{t=1}^{N}
\mathbf{1}\!\left[S_t=G_t^\ast\right],
\]
where \(N=150\).

\begin{table}[H]
\centering
\caption{Constraint applicability on 150 held-out task--scene queries. BGE-M3 retrieves the same number of constraints selected by our model for each query.}
\label{tab:constraint_applicability}
\small
\setlength{\tabcolsep}{3.5pt}
\begin{tabular}{lcccc}
\hline
Method & Recall $\uparrow$ & Precision $\uparrow$ & F1 $\uparrow$ & EMR $\uparrow$ \\
\hline
BGE-M3 Top-$n_t$ & 24.32\% & 26.11\% & 0.2519 & 0.00\% \\
Ours              & \textbf{37.48\%} & \textbf{40.23\%} & \textbf{0.3881} & \textbf{2.67\%} \\
\hline
\end{tabular}
\end{table}
As shown in Table~\ref{tab:constraint_applicability}, the proposed applicability model consistently outperforms semantic retrieval under the same rule-selection budget across all four metrics. Although absolute recall remains limited under distribution shift, the improvement indicates that the learned model captures task--environment-dependent constraint relevance beyond semantic similarity alone.

\subsection{Safety Inspector Evaluation}\label{s:4.2}

We evaluate the Safety Inspector after parameter-efficient adaptation using three VLM backbones: Qwen3-VL-30B-A3B~\cite{bai2025qwen3vl}, Qwen2.5-VL-32B~\cite{bai2025qwen25vl}, and Qwen2.5-VL-72B~\cite{bai2025qwen25vl}. For each backbone, LoRA~\cite{hu2022lora} is applied following the Inspector adaptation procedure described in Sec.~III-D, while the underlying VLM backbone remains frozen. All three Inspectors are evaluated on the same 30 unseen scenes, comprising 390 inspection samples in total. These scenes are excluded from the Inspector adaptation data.

We evaluate the Inspector using three metrics. First, \emph{Inspection Decision Accuracy} (IDA) measures the fraction of test samples for which the Inspector's \texttt{PASS}/\texttt{FAIL} decision matches the ground-truth inspection label:
\[
\mathrm{IDA}
=
\frac{N_{\mathrm{correct}}}{N_{\mathrm{test}}}.
\]

Second, \emph{Corrective Exact-Match Rate} (CEM) evaluates the quality of the corrective output for failure cases. Among ground-truth \texttt{FAIL} samples for which the Inspector correctly outputs \texttt{FAIL}, CEM measures the fraction whose complete subsequent corrective output exactly matches the expected target:
\[
\mathrm{CEM}
=
\frac{N_{\mathrm{exact}}}{N_{\mathrm{fail}}}.
\]
This metric evaluates whether the Inspector can provide the complete corrective information required by the planner after identifying a failed plan.

Finally, \emph{Rule Expansion Coverage} (REC) evaluates the Inspector's ability to reconstruct missing safety rules. Let \(m\) denote the total number of relevant safety rules involved in the 30 unseen scenes, \(n\) the number of these rules already provided to the Inspector, and \(q\) the number of missing relevant rules successfully reconstructed by the Inspector. We define
\[
\mathrm{REC}
=
\frac{q}{m-n}.
\]

\begin{table}[H]
\centering
\caption{Performance of LoRA-adapted Safety Inspectors with different VLM backbones on 30 unseen scenes comprising 390 inspection samples.}
\label{tab:inspector}
\small
\setlength{\tabcolsep}{5pt}
\begin{tabular}{lccc}
\hline
Inspector & IDA $\uparrow$ & CEM $\uparrow$ & REC $\uparrow$ \\
\hline
Qwen3-VL-30B-A3B + LoRA & 90.0\% & 35.3\% & 97.8\% \\
Qwen2.5-VL-32B + LoRA   & 96.7\% & 49.3\% & 97.8\% \\
Qwen2.5-VL-72B + LoRA   & \textbf{97.7\%} & \textbf{67.1\%} & \textbf{98.1\%} \\
\hline
\end{tabular}
\end{table}

As summarized in Table~\ref{tab:inspector}, the evaluated Inspector backbones show clear differences in both inspection decisions and corrective reasoning, with the most pronounced variation observed in CEM. In contrast, REC remains consistently high across all configurations, indicating that missing-rule reconstruction is comparatively stable across the evaluated backbones. These results suggest that generating complete corrective outputs is more challenging than making the initial \texttt{PASS}/\texttt{FAIL} decision.

\subsection{Test-Time Adaptation Evaluation}\label{s:4.3}

We next evaluate the proposed framework under a sequential test-time adaptation setting using the 50 original scenes. The experiment jointly considers the three adaptation mechanisms introduced in Sec.~III-D: updating the lightweight constraint-applicability model, adapting the Safety Inspector through LoRA, and expanding the explicit Constraint Space.For this experiment, the lightweight applicability model and Inspector LoRA parameters have not been trained or adapted on the 50 original scenes before Stage 0; only the frozen pretrained backbones are retained.

At Stage 0, the framework starts without adaptation feedback from these scenes. The 50 scenes are then processed sequentially in five stages of 10 scenes. At each stage, the current framework is first evaluated on the incoming scenes. Inspector-generated corrections are subsequently validated and used to update the constraint-applicability model, adapt the Inspector LoRA parameters, and expand the Constraint Space before proceeding to the next stage.

Consequently, Stage 0 represents the initial state before any test-time adaptation on the original scenes, while Stages 1--4 have accumulated feedback from 10, 20, 30, and 40 previously processed scenes, respectively.

We report Recall and Precision for constraint applicability using the definitions in Sec.~IV-A. In addition, Human-Verified Pass Rate (HVPR) measures the fraction of evaluated scenes for which the final plan receives \texttt{PASS} from the Safety Inspector and is subsequently verified by human review to be both safe and task-complete. The sequential results are summarized in Table~\ref{tab:tta}.

\begin{table}[H]
\centering
\caption{Sequential test-time adaptation over 50 benchmark scenes. Each stage is evaluated before feedback from its current scenes is incorporated into the adaptation process.}
\label{tab:tta}
\small
\setlength{\tabcolsep}{4pt}
\begin{tabular}{ccccc}
\hline
Stage & Adapted Scenes & Recall $\uparrow$ & Precision $\uparrow$ & HVPR $\uparrow$ \\
\hline
0 & 0  & 0.00\%  & 0.00\%  & 30.00\% \\
1 & 10 & 20.51\% & 53.33\% & 40.00\% \\
2 & 20 & 46.15\% & 40.00\% & 20.00\% \\
3 & 30 & 43.59\% & 48.57\% & 20.00\% \\
4 & 40 & 23.08\% & 60.00\% & 50.00\% \\
\hline
\end{tabular}
\end{table}

As shown in Table~\ref{tab:tta}, sequential test-time feedback moves the applicability model away from its initial zero-selection state and improves constraint prediction over deployment, although the performance is not strictly monotonic across stages due to differences in incoming-scene difficulty. HVPR is also higher in the final stage than at cold start. Inspector LoRA adaptation and Constraint Space expansion remain active throughout the experiment; however, Inspector-specific metrics are not recorded at each stage. Therefore, the reported results characterize the behavior of the complete adaptation framework rather than isolating the stage-wise contribution of Inspector adaptation.
\subsection{Real-World Robot Experiments}\label{s:4.4}

We further conduct representative real-world experiments on a physical robotic platform to examine whether the safety constraints selected by the proposed framework can be reflected in executable manipulation behaviors. The high-level plans are expressed using the predefined manipulation primitives described in Sec.~III-E and are grounded in the physical scene only after passing safety inspection.

\begin{figure}[H]
    \centering    \includegraphics[width=\linewidth]{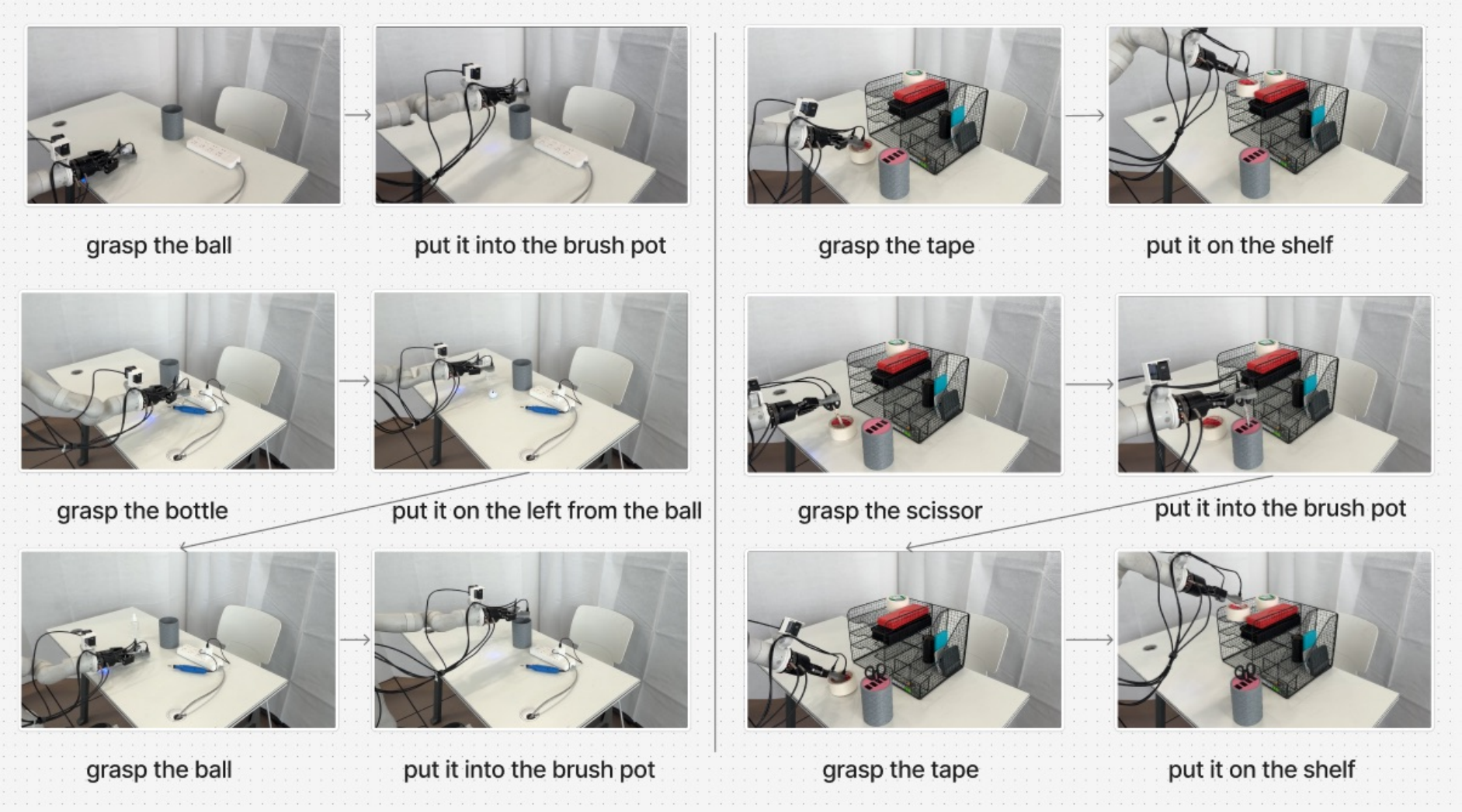}    \caption{\textbf{Representative real-world robot experiments.}
Direct task plans are compared with safety-aware plans generated after incorporating scene-dependent constraints. For the ball-placement task, the safety-aware plan first relocates the nearby bottle before manipulating the ball. For the tape-placement task, the nearby scissors are first removed from the manipulation region. In both cases, the added safety operations are expressed using the same predefined \texttt{Grasp} and \texttt{Put} primitives as the original task.}
    \label{fig:3}
\end{figure}
Figure~\ref{fig:3} presents two representative comparisons. In the first task, the robot is instructed to grasp a ball and place it into a brush pot. Without additional scene-dependent safety reasoning, the direct plan consists only of
\emph{Grasp(ball)} followed by \emph{Put(ball, brush pot)}. When the nearby bottle is identified as a potential obstruction or hazard to the manipulation, the safety-aware plan first inserts \emph{Grasp(bottle)} and \emph{Put(bottle, away from ball)}, and then resumes the original task.

A similar behavior is observed in the second task, where the robot is instructed to place a tape roll onto a shelf. The direct plan contains only \emph{Grasp(tape)} and \emph{Put(tape, shelf)}. With the applicable safety constraint, the planner first removes the nearby scissors from the manipulation region using \emph{Grasp(scissors)} and \emph{Put(scissors, brush pot)}, before executing the requested tape-placement task.

These examples demonstrate that the proposed framework preserves the original task objective while modifying the primitive sequence when additional scene-dependent safety requirements are identified. Importantly, the inserted safety operations use the same predefined manipulation primitives as the original task, allowing the resulting plans to remain compatible with the existing grounding and execution pipeline.
\section{DISCUSSION AND LIMITATIONS}\label{s:5}

\textbf{Inspector adaptation.}
The framework adapts the Safety Inspector through LoRA together with constraint-applicability adaptation and Constraint Space expansion. Sec.~IV-B shows that decision accuracy and rule-expansion coverage remain high across the evaluated backbones, while exact corrective generation is more sensitive to model capacity. During sequential test-time adaptation, Inspector adaptation remains active, although Inspector-specific metrics are not recorded at every stage; the reported results therefore reflect the behavior of the complete adaptation framework rather than the isolated contribution of Inspector adaptation.

\textbf{Limitations.}
The proposed framework operates at the behavior-planning level and does not replace low-level collision avoidance, motion-level safety control, or formal verification. The current evaluation is limited to tabletop manipulation and textual safety constraints, and the long-term stability of continual Inspector adaptation under longer deployment sequences has not yet been systematically evaluated. Future work will consider temporal constraints, richer physical interactions, longer-horizon manipulation tasks, and longer-term adaptation.
\section{CONCLUSION}\label{s:6}

We presented a safety-aware robotic behavior planning framework built around an explicit and extensible Constraint Space. A query-conditioned model selects task--environment-relevant constraints, while a multimodal Safety Inspector performs closed-loop verification, constraint reconstruction, and corrective re-planning. Validated feedback further supports constraint-applicability adaptation, LoRA-based Inspector adaptation, and explicit safety-knowledge expansion. Experiments on constraint selection, Safety Inspector performance, sequential test-time adaptation, and real-world manipulation demonstrate the effectiveness of integrating explicit safety knowledge with deployment-time corrective feedback.

\bibliographystyle{IEEEtran}
\bibliography{references}

\end{document}